\documentclass[10pt,twocolumn,letterpaper]{article}

\usepackage[pagenumbers]{iccv} % To force page numbers, e.g. for an arXiv version

\usepackage[accsupp]{axessibility}  % Improves PDF readability for those with disabilities.
\usepackage{comment}
\usepackage{graphicx}
\usepackage{booktabs}
\usepackage{multirow}
\usepackage{adjustbox}
\usepackage{makecell}
\newcommand{\method}{MotionShot\xspace}

\definecolor{iccvblue}{rgb}{0.21,0.49,0.74}
\usepackage[pagebackref,breaklinks,colorlinks,allcolors=iccvblue]{hyperref}

\def\paperID{5124} % *** Enter the Paper ID here
\def\confName{ICCV}
\def\confYear{2025}

\title{MotionShot: Adaptive Motion Transfer across Arbitrary Objects for Text-to-Video Generation
}

\author{
Yanchen Liu$^{1}$,
Yanan Sun$^{3}$\textsuperscript{\dag},
Zhening Xing$^{3}$,
Junyao Gao$^{4}$,
Kai Chen$^{3}$,
Wenjie Pei$^{1,2}$\textsuperscript{\dag} \\
$^{1}$Harbin Institute of Technology (Shenzhen), \\
$^{2}$Peng Cheng Laboratory,
$^{3}$Shanghai AI Laboratory,
$^{4}$Tongji University \\
{\tt\small lyc.codeplayer@outlook.com, wenjiecoder@outlook.com, junyaogao@tongji.edu.cn,} \\
{\tt\small \{sunyanan, xingzhening, chenkai\}@pjlab.org.cn}
}

\begin{document}

\twocolumn[{%
\renewcommand\twocolumn[1][]{#1}%s
\maketitle
\vspace{-3em}
\begin{center}
    \centering
    \captionsetup{type=figure}
    \includegraphics[width=\textwidth]{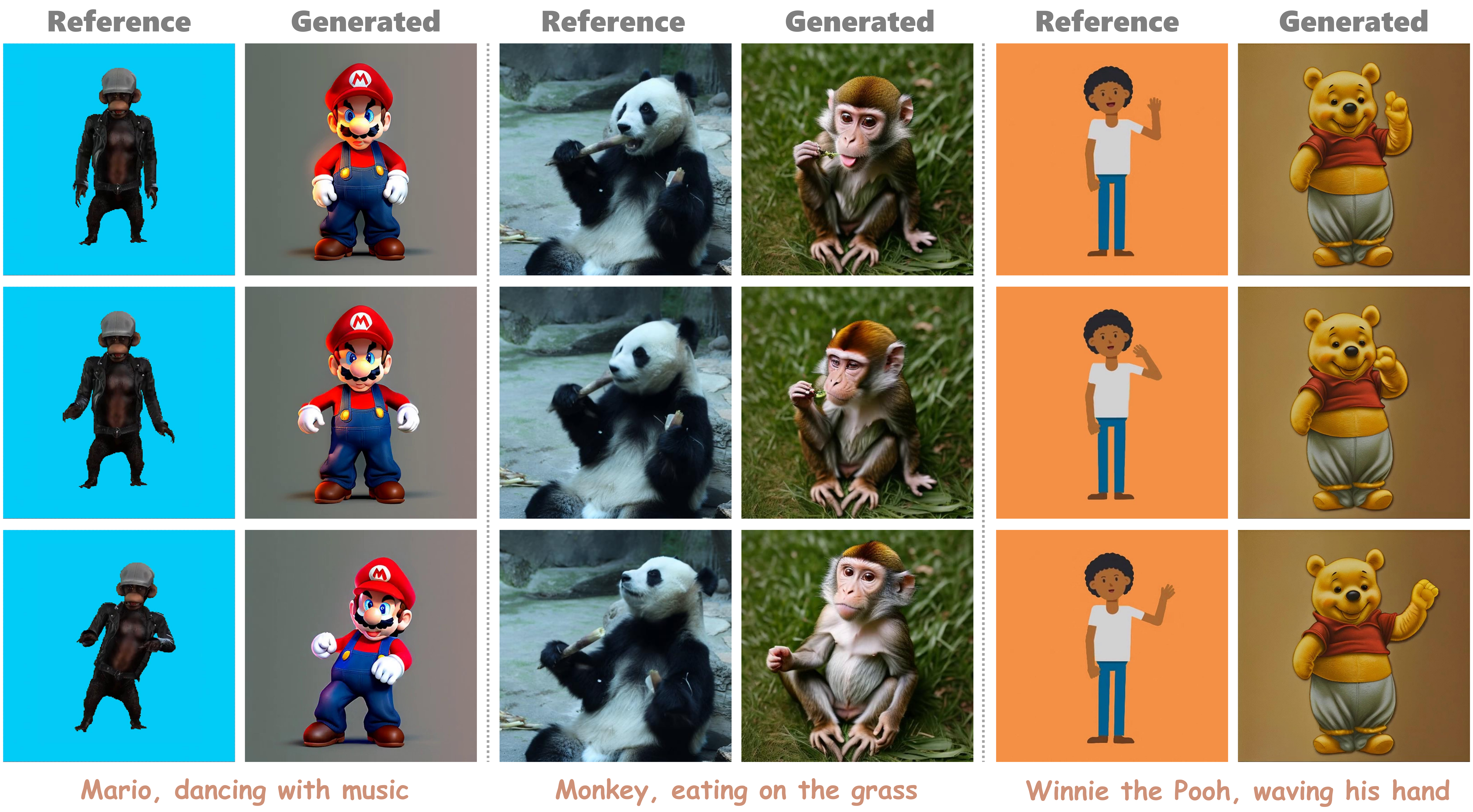}
    \vspace{-2em}
\captionof{figure}{\textbf{Visualization results of our \method}. Given any reference video, \method can effectively transfer motion across arbitrary objects in text-to-video generation. Notably, when the reference and target objects have distinct appearances and structures—such as an anime boy and a Winnie bear—\method demonstrates remarkable adaptive motion transfer capabilities.}
\label{fig:teaser}
    % \vspace{0.3cm}
\end{center}}]
%Our method is a training-free approach that enables the transfer of motion across arbitrary objects in text-to-video generation. This capability persists even when the reference and target objects have distinct appearances and structures. 

\makeatletter
\renewcommand{\thefootnote}{\dag}
\makeatother
\footnotetext[1]{Corresponding author.}
\begin{abstract}
%Existing text-to-video methods struggle to transfer motion smoothly from a reference object to a target object with significant differences in appearance or structure between them. To address this, we introduce \method, a training-free framework capable of transferring motions reasonably across objects with substantial appearance and structure disparities, keeping semantic as well as topological alignment in motion while preserving the coherence in object appearance. \method generates motion-controllable videos through temporal attention guidance that is curated based on high-level and low-level motion alignment. We achieve the two-level alignment by establishing semantic correspondence and retargeting the reference object shape to that of the target object respectively. Experiments demonstrate \method's superior performance in motion fidelity and structural coherence, especially when transferring motion between objects with large appearance and structural differences.

\vspace{-1em}
Existing text-to-video methods struggle to transfer motion smoothly from a reference object to a target object with significant differences in appearance or structure between them. To address this challenge, we introduce \method, a training-free framework capable of parsing reference-target correspondences in a fine-grained manner, thereby achieving high-fidelity motion transfer while preserving coherence in appearance. To be specific, \method first performs semantic feature matching to ensure high-level alignments between the reference and target objects. It then further establishes low-level morphological alignments through reference-to-target shape retargeting. By encoding motion with temporal attention, our \method can coherently transfer motion across objects, even in the presence of significant appearance and structure disparities, demonstrated by extensive experiments. The project page is
available at: https://motionshot.github.io/.
%generates motion-controllable videos through temporal attention guidance that is curated based on high-level and low-level motion alignment. We achieve the two-level alignment by establishing semantic correspondence and retargeting the reference object shape to that of the target object respectively. 
%Extensive experiments demonstrate \method's superior performance in motion fidelity and structural coherence, especially when transferring motion between objects with large appearance and structural disparities.

% and low-level  representation mapping semantic topological alignment-based  with three key components: 
% 1) semantic keypoint matching to establish structural correspondences; 2) shape retargeting to align motion with the target object's structure; and 3) attention-guided video generation to ensure motion fidelity and structural coherence. Experiments demonstrate \method's superior performance in motion fidelity and structural coherence, especially when transferring motion between objects with large appearance and structural differences.
\end{abstract}

\vspace{-1em}
\section{Introduction}
\label{sec:intro}

Recent advances in diffusion models \cite{ho2020denoising, dhariwal2021diffusion, song2020denoising} have significantly propelled the progress of video generation \cite{singer2022make, li2023videogen, yin2023nuwa, guo2023animatediff}. Although existing methods can produce high-quality videos guided by text prompts, achieving precise motion customization---where generated videos adhere to specific motion patterns from user-provided reference videos---remains particularly challenging, especially for \textbf{arbitrary reference-target object pairs} with \textbf{significant appearance differences}.
% maintain specific motion patterns according to user-provided reference videos, remains particularly challenging.

Existing motion transfer methods primarily focus on developing effective motion descriptors. For example, one line of research~\cite{ma2024followpose, hu2024animate} utilizes landmark sequences as motion descriptors for transferring motion between reference-target pairs with similar appearances. However, this approach cannot be easily generalized to arbitrary objects, as predefining landmarks for all objects proves to be challenging. Another approach~\cite{ma2024follow} typically extracts learned spatial-temporal features from reference videos as motion descriptors. Unfortunately, the inherent entanglement of motion and appearance in latent representations creates a critical bottleneck, leading to unintended leakage of reference appearance details. Recent studies have turned to alternative motion cues as intermediate motion descriptors, including depth or edge maps~\cite{wang2023videocomposer,chen2023control,xing2024make,esser2023structure,guo2024sparsectrl}, sparse optical flow or trajectories~\cite{yin2023dragnuwa,niu2024mofa,wang2024motionctrl,lei2024animateanything}. While these methods excel at transferring motion between objects with minor appearance differences, they often struggle with objects that have substantial appearance discrepancies, as they do not account for \textbf{region-level semantic correspondence} and \textbf{pixel-level structural correspondence}.
In this work, we introduce \textbf{\method}, a new training-free motion transfer framework capable of accurately transferring motion information to a target object without requiring additional training, even when there are considerable differences in appearance and structure, as illustrated in~\cref{fig:teaser}. Our \method directs video generation to adhere to the desired motion using temporal attention guidance, eliminating the need for labor-intensive large-scale data collection. However, attention guidance based on positional alignment becomes less effective for objects with substantial appearance differences. To tackle this issue, we propose a novel two-level motion alignment strategy---\textit{high-level semantic motion alignment} and \textit{low-level structural motion alignment}---to create adaptive temporal attention guidance for arbitrary object pairs. 

Specifically, the high-level motion alignment establishes semantic correspondence automatically between reference and target objects. This correspondence is determined through semantic feature matching between two keypoint sets, which are sampled in a structure-aware manner from both the reference and target objects. Relying solely on high-level motion alignment may lead to discontinuities in temporal attention guidance. We further enhance the motion alignment with low-level structural mapping, achieved through Thin Plate Spline-based shape warping. This approach ensures more precise motion control while maintaining structural alignment with the target object.
By integrating our two-level motion alignment, the attention-guided video generation model enables motion transfer that faithfully follows the reference motion while naturally fitting the structure of the target subject. \method is the \textit{first} framework to explicitly model both high-level and low-level motion alignment. Overall, our main contributions are manifold:

\begin{itemize}
    \item We introduce \method, a novel training-free motion transfer framework that facilitates precise motion adaptation, even when there are substantial differences in appearance and structure between the reference and target objects.
    \item We develop an unique two-level motion alignment strategy that combines semantic and structural alignment to establish correspondence between reference-target pairs, allowing for adherence to the reference object's motion while preserving the appearance of the target object.
    \item \method demonstrates superior performance compared to existing methods in both motion fidelity and structural coherence, particularly in scenarios where there are significant appearance and structural discrepancies between the reference and target objects.
\end{itemize}

\section{Related Work}
\label{sec:related}

\subsection{Text-to-Video Diffusion Models}
With the significant development of diffusion models \cite{gao2024styleshot,nichol2021glide,ramesh2022hierarchical,rombach2022high} in generating high-quality images, recent methods \cite{guo2024sparsectrl, jiang2023text2performer, wang2024lavie} emerge the diffusion model as a leading technology in text-to-video generation.
Specifically, Video Diffusion Model \cite{ho2022video} leverages an innovative 3D U-Net \cite{ronneberger2015u} architecture to generate temporally consistent videos, and \cite{zhou2022magicvideo, he2022latent} extend this approach into the latent space to tackle data scarcity, complex temporal dynamics, and high computational costs.
% by innovating the 3D U-Net \cite{ronneberger2015u} architecture, Video Diffusion Model \cite{ho2022video} enables the model to generate temporally consistent videos. Based on this, \cite{zhou2022magicvideo, he2022latent} build 3D U-Net on the latent space to tackle the data scarcity, complex temporal dynamic and high computational costs.
Moreover, \cite{chen2023videocrafter1, xing2024dynamicrafter, blattmann2023stable, zhang2024show} construct the well-organized text-video dataset and propose the decouple strategy to enhance the temporal-spatial coherence in video diffusion models.
Another trend, \cite{singer2022make, li2023videogen, yin2023nuwa, guo2023animatediff} extends text-to-image diffusion models' ability to generate video by fine-tuning the extra temporal layers in the pre-train text-to-image diffusion model.
Models like \cite{khachatryan2023text2video, wu2023tune} employ special-designed frame-attention mechanism to enable the zero-shot text-to-video generation.
Recently, DiT \cite{peebles2023scalable} has been integrated into text-to-video generation, leading to extensive research efforts \cite{ma2024latte, gupta2024photorealistic} and notable productions \cite{zheng2024open, kong2025hunyuanvideosystematicframeworklarge, yang2024cogvideox}, which enhances temporal consistency and motion coherence in generated videos, enabling high-fidelity synthesis from textual descriptions.
% In our work, \method is built upon AnimateDiff \cite{guo2023animatediff}, aiming to generate high-quality videos that align with both the textual content and the motion of the reference video. 

\begin{figure*}[htbp]
\vspace{-1em}
  \centering
  \includegraphics[width=\textwidth]{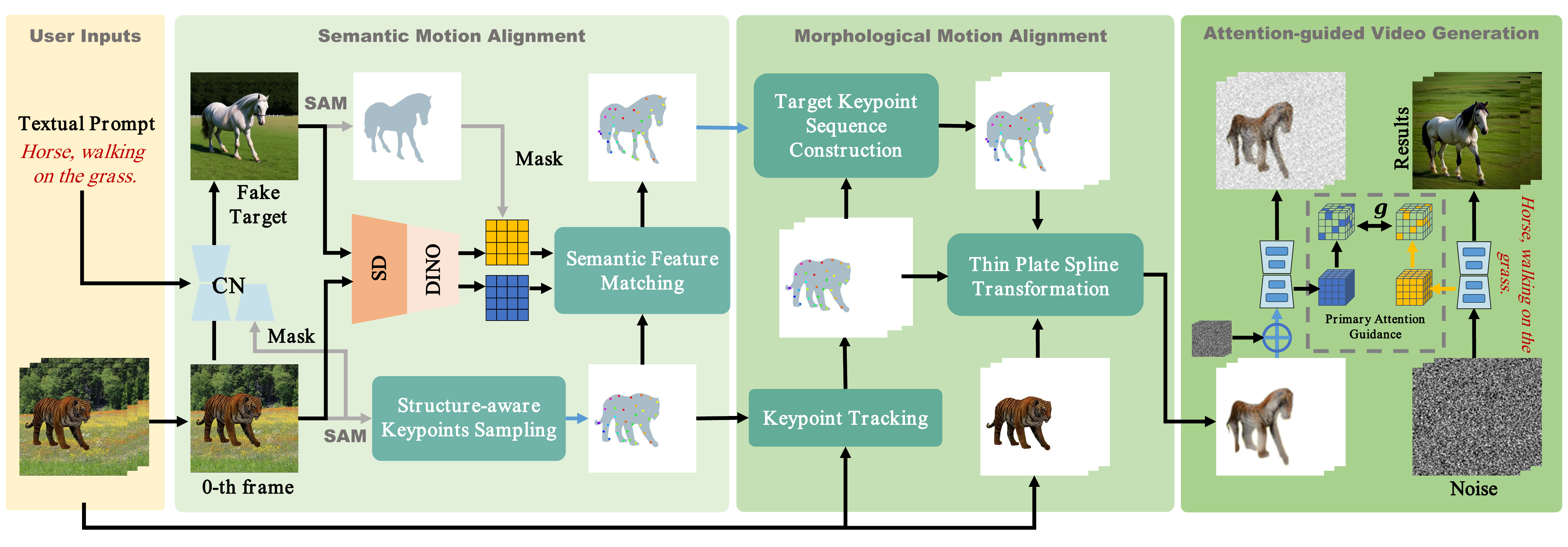}
  \caption{
  \textbf{The architecture of \method}, a training-free motion transfer method capable of handling reference-target object pairs with substantial appearance difference. A novel two-level motion alignment strategy, high-level semantic motion alignment as well as low-level morphological motion alignment, is introduced to establish the adaptive temporal attention guidance, leading to effective motion transfer. 
  % consists of three components. % In Semantic Keypoint Matching module, we first build semantic correspondence between the reference object and a fake target object which is derived from a segmentation ControlNet with the user-provided textual prompt. The established correspondence will guide the transformation from the reference shape sequence to the target shape sequence binding with the same motion information in our shape retargeting module. Afterwards, the retargted shape sequence provides coherent structures with the object described in the textual prompt, thus leading to effective motion transfer. 
  }
  \label{fig:framework}
  \vspace{-1em}
\end{figure*}

\subsection{Motion Transfer}
% Motion transfer involves generating videos that preserve the motion characteristics of the reference video, including direction, speed, and posture, while adapting the dynamic subject to match the form described in the text prompt. 

\begin{comment}
    Motion transfer aims to generate videos that inherit the reference video's motion attributes (e.g., direction, speed, posture) while adapting the subject to match the appearance and style specified by the text prompt.
Existing methods adopt different strategies for motion transfer. Some approaches \cite{ma2024follow, gao2025faceshot, hu2024animate, ma2024followpose} use landmark sequences (e.g., facial keypoints or skeleton maps) to accurately capture and transfer motion trajectories, but defining landmarks for arbitrary objects is challenging. Other methods adopt strategies from image synthesis\cite{zhang2023adding,zhao2023uni}, leveraging features such as depth maps and edge maps \cite{wang2023videocomposer,chen2023control,xing2024make,esser2023structure,guo2024sparsectrl} extracted from reference videos to guide motion. However, overly dense conditioning can reveal excessive structural information, compromising transfer quality. Alternatively, some studies explore using trajectory information or sparse optical flow for control\cite{yin2023dragnuwa,niu2024mofa,wang2024motionctrl,lei2024animateanything}, reducing the leakage of appearance details, but these approaches limit the scope of transferable motion information. 
\end{comment}

Motion transfer aims to generate videos that inherit the motion attributes (e.g., direction, speed, posture) of a reference video while adapting the subject's appearance and style based on a text prompt. 
Some approaches train the text-to-video generation model with external conditions, including keypoints\cite{ma2024followpose,ma2024follow,hu2024animate}, depth maps\cite{guo2024sparsectrl,wang2023videocomposer,chen2023control,esser2023structure}, edge maps\cite{xing2024make,wang2023videocomposer}, sparse optical flow or trajectory\cite{yin2023dragnuwa,niu2024mofa,wang2024motionctrl,lei2024animateanything,wang2024motion}
Some studies have attempted to train or fine-tune the model with video contains specific motion concept\cite{jeong2024vmc,zhao2024motiondirector}.
However, such training-based methods often domain-specific and require extensive data collection.
Some research has shifted towards training-free methods that utilize attention mechanisms \cite{ling2024motionclone, meral2024motionflow, gao2025faceshot} or motion consistency loss \cite{zhang2025training} to improve generalization. However, these methods perform well with objects that share similar appearances and structures but often struggle with distinct objects. In this paper, we introduce \method, a training-free motion transfer framework which can effectively transfer motion information across a variety of target objects.

\subsection{Attention-Based Guidance}
% Recent studies \cite{hedlin2024unsupervised, luo2023diffusion, tang2023emergent, zhang2023tale} have demonstrated that the features and attention mechanisms in diffusion models encapsulate a wealth of information, exhibiting remarkable generalization capabilities. These characteristics enable diffusion models to precisely capture complex visual concepts, showcasing strong performance in image and video generation and editing tasks.
% Many studies have focused on leveraging attention mechanisms to enhance visual generation. 
% For instance, \cite{chefer2023attend, meral2024conform} introduced cross-attention constraints during inference to effectively refine latent features in diffusion models, mitigating issues such as subject omission and incorrect attribute binding in image generation. Additionally, \cite{tumanyan2023plug} employed self-attention mechanisms to impose semantic layout constraints on images, enabling more precise control over image editing and improving layout consistency and editability. 
Recent studies \cite{hedlin2024unsupervised, luo2023diffusion, tang2023emergent, zhang2023tale,gao2023similarity} have shown that the features and attention mechanisms in diffusion models encapsulate extensive information and demonstrate strong generalization capabilities. This allows diffusion models to effectively capture complex visual concepts, excelling in content generation and editing tasks. For instance, \cite{chefer2023attend, meral2024conform} introduce cross-attention constraints to refine latent features, addressing issues like subject omission and attribute misbinding in image generation. Additionally, \cite{tumanyan2023plug} uses self-attention mechanisms to impose semantic layout constraints, enhancing control, layout consistency, and editability in images.

In motion transfer tasks, attention mechanisms are commonly employed for motion extraction and control. For example, \cite{meral2024motionflow} uses cross-attention features to extract key motion information, guiding the spatial dynamics of target subjects across frames. Additionally, \cite{ling2024motionclone} examines temporal attention layers, demonstrating their capacity to encode global motion dynamics and represent motion with sparse temporal attention weights, aiding in motion transfer. However, these methods face motion incompatibility issue due to the strong coupling between motion and structure when the target and reference objects differ significantly.

% In motion transfer tasks, attention mechanisms are also widely used for motion extraction and control. \cite{meral2024motionflow} utilized cross-attention features to extract primary motion information, effectively guiding the spatial dynamics of target subjects in different frames. Furthermore, \cite{ling2024motionclone} explored the role of temporal attention layers, revealing their ability to encode global motion dynamics and represent motion through sparse temporal attention weights, thereby facilitating motion transfer. Although these methods achieve motion transfer through attention mechanisms, they struggle with motion incompatibility due to the strong coupling between motion and spatial structure when the target and reference objects differ significantly. 
% To address this, we propose a novel attention-guided video generation framework based on semantic matching and shape retargeting, which adapts motion information in temporal attention layer to align with the target object's structure, ensuring natural and coherent video generation.

\subsection{Motion Retargeting}
Motion retargeting is a technique to adapt existing motion from a reference object to a target object with different appearance and structures, which is an essential step in motion transfer. Early works formulate motion retargeting as a constrained optimization problem~\cite{Choi2000OnlineMR, Gleicher1998RetargettingMT, Lee1999AHA, Popovic1999PhysicallyBM}. These methods usually require a tedious and time-consuming process of designing constraints tailored to specific motion sequences. With the advent of deep learning, researchers have increasingly focused on learning-based motion retargeting methods ~\cite{Aberman2020SkeletonawareNF,Hu2023PoseAwareAN,Lim2019PMnetLO,Villegas2018NeuralKN, Villegas2021ContactAwareRO,Zhang2023SkinnedMR} in recent years. However, most existing methods are specifically designed for human motion retargeting, focusing primarily on joint-relative relationships while often overlooking high-level semantic information. Furthermore, generalizing motion retargeting to arbitrary objects presents a significant challenge, as it is an ill-posed problem that lacks prior information.

\section{Method}
\label{sec:method}
The framework of \method is depicted in \cref{fig:framework}. We first introduce \method in \cref{sec:3.1}, then elaborate our motion transfer framework with the novel semantic and morphological motion alignment in~\cref{sec:3.2,sec:3.3,sec:3.4}.

\subsection{Overview}
\label{sec:3.1}
% In the keypoint matching stage, we utilize the 
% The Stable Diffusion model~\cite{rombach2022high} is a powerful text-to-image (T2I) generative model designed for creating high-quality images from textual descriptions, leveraging advanced diffusion processes and latent space representations to enable control over the generated content. Similarly, video diffusion models generate vivid videos from user-provided textual prompts. In video diffusion models, the transition from Text-to-Image to Text-to-Video (T2V) is primarily achieved by integrating a temporal transformer module into the foundational block. This module employs a frame-level self-attention mechanism along the temporal dimension  to model inter-frame correlations, thereby preserving the temporal continuity essential for dynamic video content.

% Specifically, the input is a spatio-temporal feature tensor with an initial shape of ${C \times F \times H \times W} $, where \( C \), \( F \), \( H \), and \( W \) represent the number of channels, frames, height, and width, respectively. 
% For simplicity, we omit the batch size dimension (which equals one) in our notation.
% The tensor is then transformed into a feature tensor of shape $(H \times W) \times F \times C$. This module employs a frame-level self-attention mechanism, utilizing the temporal attention mechanism along the \( F \) dimension (corresponding to video frames) to model inter-frame correlations, thereby preserving the temporal continuity essential for dynamic video content.

In text-to-video generation, textual prompts generally offer video-level descriptions for video generation, lacking fine-grained control over object motion. In practical scenarios where users need precise control over object movement, they often provide a reference video that demonstrates the desired motion. This process, which involves transferring the motion depicted in the reference video to the target generated object, is known as motion transfer. % In this work, we develop our motion transfer framework in a training-free manner by manipulating temporal attention guidance.

Achieving motion retargeting between arbitrary reference and target object pairs in text-to-video generation is challenging due to the complexity of establishing semantic correspondence. Structural variations make predefined correspondences, like skeleton keypoints, impractical, and the actual shape of the target object remains unknown until generated. To tackle this, we first create a fake target object based on the user-provided textual prompt, which helps establish semantic correspondence for high-level motion alignment with the reference object. We then refine this alignment through shape warping at a lower level. The motion guidance from these two levels ensures that the generated videos maintain semantic and morphological consistency with the reference object's motion while achieving a natural appearance aligned with the textual prompt.

% Our method comprises three key modules: 1) Semantic Motion Alignment, 2) Structural Motion Alignment, and 3) Attention-Guided Video Generation.

% Based on this semantic correspondence, we can achieve motion retargeting and guide video generation using the crafted motion information. 

% We propose a novel training-free motion transfer framework that seamlessly integrates three key components for high-quality motion synthesis. 
% First, the semantic keypoint matching module establishes precise correspondences between the keypoints of the reference video and those of the target subject based on semantic consistency. Next, the shape retargeting module generates the target keypoint sequence and applies the Thin Plate Spline (TPS) transformation to warp video frames, aligning the motion information of the reference video with the structural appearance of the target subject. Finally, the attention-guided video generation model synthesizes the final video, preserving motion fidelity and structural integrity to achieve high-quality outputs.

\subsection{Semantic Motion Alignment}\label{sec:3.2}

\noindent{\textbf{Fake target object generation.}} Naturally, we can generate a fake target object directly using a well-pretrained text-to-image model according to the user-provided textual prompt and then establish semantic correspondence based on the reference object and fake target object. However, we find that when two objects have distinct initial poses, the motion transfer becomes unstable. To address this issue, our fake target object generation process also takes the first frame of the reference video as input, providing the initial pose information for the generated target object.

\begin{figure}
\vspace{-1.5em}
    \centering
    \includegraphics[width=1.0\linewidth]{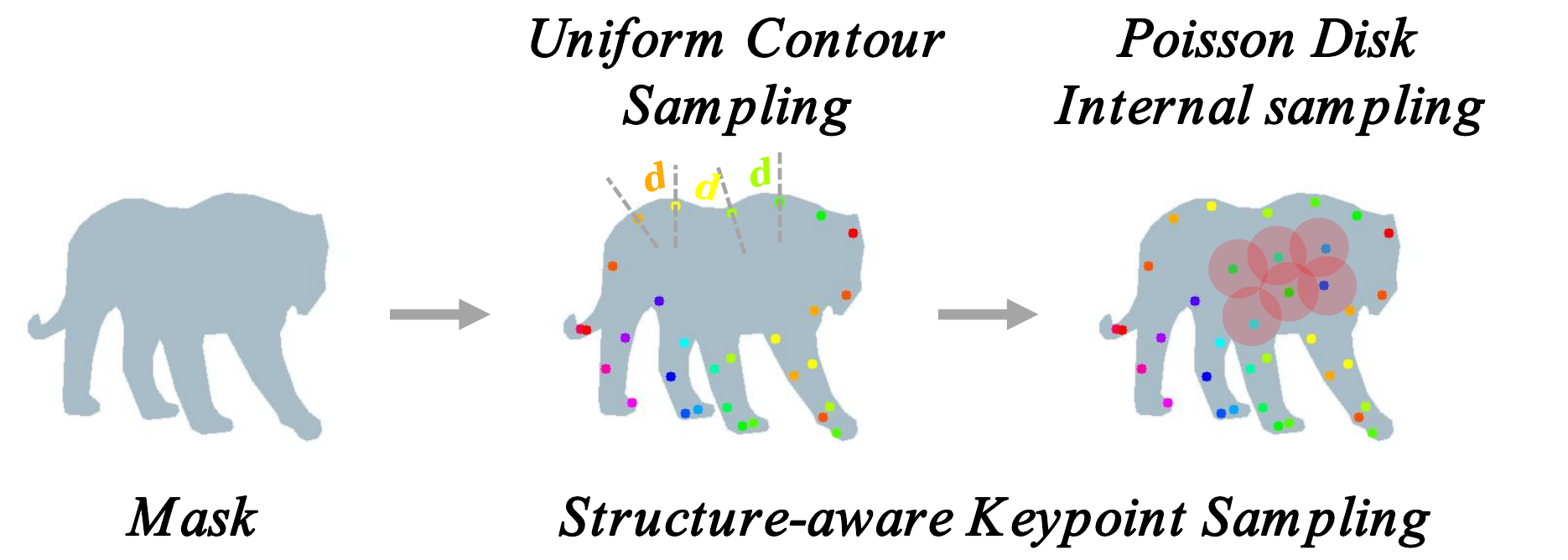}
    \vspace{-1.5em}
    \caption{\textbf{Structure-aware keypoint sampling} consisting of uniform contour sampling and Poisson disk internal sampling.}
    \label{fig:keypoint_sampling}
    \vspace{-1.5em}
\end{figure}

Specifically, we utilize StableDiffusion-ControlNet-segmentation~\cite{zhang2023adding} as the text-to-image model, inputting a degraded segmentation map of the reference object along with textual prompt. We use a degraded segmentation map because an accurate map would reveal the structure of the reference object, whereas we only need a coarse hint of the initial pose. To further mitigate the negative impact of the reference object shape, we set the segmentation condition weight to a small value, ensuring that the textual prompts dominate the generation process. Ultimately, we obtain a fake target object that meets user requirements while sharing a similar initial pose with the reference object.

\vspace{0.5em}
\noindent{\textbf{Structure-aware keypoint sampling.}} After obtaining a fake target object, we establish semantic correspondence between the reference and target images through keypoint feature matching. Matched keypoints serve as anchors for motion retargeting. However, determining the location and number of keypoints for correspondence matching is challenging. Pre-defining keypoints for arbitrary objects is impractical. While open-world keypoint detection offers a viable solution, the generated keypoints are too sparse, making motion retargeting difficult. Thus, we propose a structure-aware keypoint sampling strategy including \textit{uniform contour sampling} and \textit{Poisson disk internal sampling}.

\begin{figure}
    \centering
    \vspace{-1.5em}
    \includegraphics[width=1.0\linewidth]{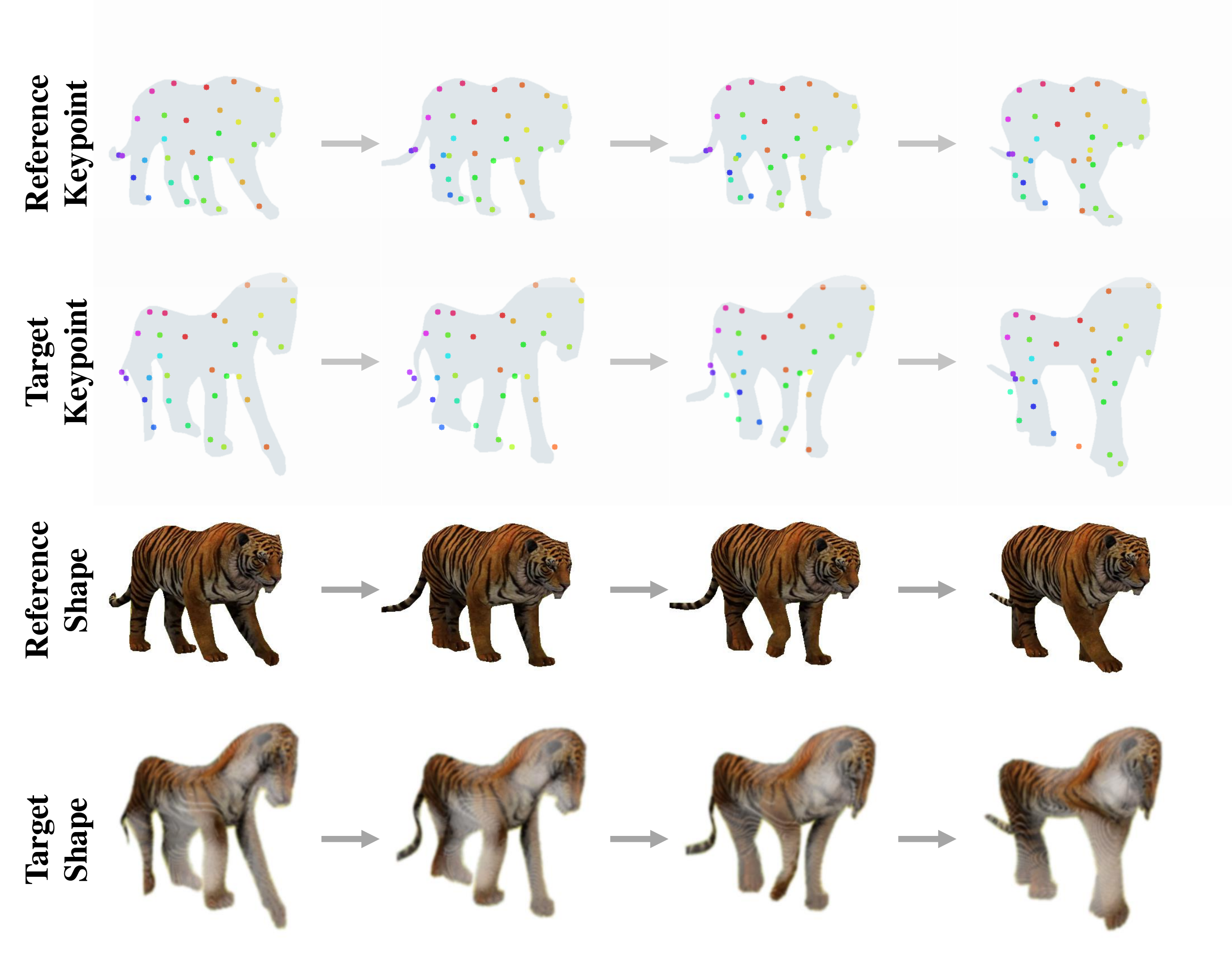}
    \caption{\textbf{TPS-based shape warping} transfers the motion of the reference object while preserving the structure of the target object.}
    \label{fig:shape-adaptation}
    \vspace{-1.5em}
\end{figure}
\vspace{0.5em}

Specifically, we first segment the reference object from the first frame $I_\text{ref}$  of the reference video and the target object from the generated fake image $I_\text{fake}$ using SAM~\cite{Kirillov2023SegmentA}. Then, we sample a set of keypoints along the contour of the reference segmentation map at uniform intervals $d$. Subsequently, we employ Poisson disk sampling to sample additional keypoints within the interior of the reference segmentation map. These $m$ keypoints collectively form the reference object keypoint set, denoted as $K_\text{ref}^0$. An illustration of this process is shown in Figure~\ref{fig:keypoint_sampling}.

Correspondingly, we identify the matching keypoint locations on the target object through semantic feature matching to construct the target keypoint set $K_\text{tar}^0$. This approach ensures that the keypoints are scatteredly distributed across different regions of the object while maintaining semantic correspondence. As a result, it achieves a region-level semantic alignment between the reference and target objects, effectively preserving the spatial consistency of key regions. We elaborate the semantic feature matching as follows.

\begin{comment}
\begin{figure}
    \centering
    \includegraphics[width=1.0\linewidth]{imgs/feature_matching.pdf}
    \caption{Semantic feature matching.}
    \label{fig:feature_matching}
\end{figure}
\end{comment}

\vspace{0.5em}
\noindent{\textbf{Semantic feature matching.}} We take both low-level and high-level features into consideration when performing semantic feature matching. 
Previous studies \cite{hedlin2024unsupervised,luo2023diffusion,tang2023emergent,zhang2023tale, zhang2023tale} have demonstrated that diffusion features exhibit strong semantic correspondences and generalization capabilities. That is, feature matching can be used to map pixels from the reference image $I_\text{ref}$ to the most similar pixels in the target image $I_\text{tar}$.  \cite{zhang2023tale} further highlights that stable diffusion features primarily focus on low-level spatial information, ensuring spatial coherence in correspondences. 
% particularly in the absence of strong texture signals. 
In contrast, features extracted from DINO \cite{amir2021deep} capture high-level semantic information and excel at obtaining sparse yet precise matches. Since these two types of features complement each other, combining them can significantly enhance the accuracy of semantic correspondence establishment.

We acquire the diffusion features from Stable Diffusion model~\cite{rombach2022high} following~\cite{zhang2023tale}. Simply put, we employ the DDIM inversion process for $I_\text{ref}$ and $I_\text{fake}$, take the diffusion features $f_\text{ref}$ and $f_\text{tar}$ from selected U-Net layers, and then perform principal component analysis \cite{mackiewicz1993principal} on the concatenation of $f_\text{ref}^\text{sd}$ and $f_\text{tar}^\text{sd}$ (layer index is ignored for simplicity) to obtain the reduced each layer’s dimension-reduced features, which are upsampled to the same resolution to form the final diffusion feature $\tilde{f}_\text{ref}^\text{sd}$ and $\tilde{f}_\text{tar}^\text{sd}$. The concatenation is performed along the spatial dimension before PCA to project two images into a common subspace, enabling subsequent feature alignments between the two images.

For DINO features, we refer to the token features from layer 11 of DINOv2 \cite{oquab2023dinov2} as $f^{\text{dino}}$. Finally, the semantic feature $f^\text{s}$ is the concatenation of the $L_2$-normalized $\tilde{f}^\text{sd}$ and $f^\text{dino}$. The similarity is computed as the following equation,
% through the distance matrix of $f_\text{ref}^\text{s}$ and $f_\text{tar}^\text{s}$ as formulated, 
\begin{equation}
    Sim(i, j) = -\| f_{\text{tar}}^s(i) - f_{\text{ref}}^s(j) \|_2,
\end{equation}  
where $i$ is referred to the pixel index in the target image while $j$ is the position of the $j$-th keypoint in the reference image. For each keypoint, we takes the most similar pixel as the matched target point.

\subsection{Morphological Motion Alignment}\label{sec:3.3}
We further refine the high-level motion alignment through low-level morphological motion alignment, where two key steps are involved: \textit{target keypoint sequence construction} and \textit{TPS-based shape warping}.
% adapt the reference motion from the reference object to the target object while preserving its structural integrity. 

\vspace{0.5em}
\noindent{\textbf{Target keypoint sequence construction.}} 
While it is possible to perform semantic motion alignment on a frame-by-frame basis to create a target keypoint sequence that captures the desired motion information, this approach often results in flickering. To overcome this challenge, we construct the target keypoint sequence using pixel tracking and motion shifts. We begin by tracking the movements of sampled keypoints across successive frames in the reference video with CoTracker3~\cite{karaev2024cotracker3}, resulting in the reference keypoint sequence \( \mathbf{K}_{\text{ref}} = [ K_{\text{ref}}^0, K_{\text{ref}}^1, \dots, K_{\text{ref}}^{F-1} ]\). Given the initial target keypoint set $ K_{\text{tar}}^0$ and the reference keypoint sequence $\mathbf{K}_{\text{ref}}$, we then generate the corresponding target keypoint sequence $\mathbf{K}_{\text{tar}}$ by computing the delta motion between neighboring frames. Generally, we first compute a global delta motion for the whole keypoint set and then refine each point coordinate with local delta motion. 

Specially, we estimate the global motion for keypoint set by fitting an ellipse characterized by a center $O$ and orientation $\Theta$, and computing the delta motion as the rotation shift $\Delta\Theta$ and the relative center shift $\Delta O$ between two neighboring reference keypoint sets. 
Subsequently, we determine the keypoint set for the target frame at timestamp $t$ by applying the rotation and center shift transformations as~\cref{eq:transform},
\begin{align}\label{eq:transform}
     K^t_{\text{tar}} = \mathcal{S}(\mathcal{R}(K^0_\text{tar}, \Delta\Theta^t), \Delta O^t),
\end{align}
where $\mathcal{R}$ and $\mathcal{S}$ denote rotation and shift operation, respectively. 

% As known, a point set has diameter $d$, which is the segment connecting the most distant two points in the point set. We name the angle between the diameter and x-axis as the rotation angle $\theta$ and the diameter midpoint as the center $c$ of the point set. 
% For the keypoint sets of two neighboring source frames, we define the delta motion as the rotation change $\Delta\theta$ and the relative center shift $\Delta \hat{c} = \Delta c/|d|$. 
% we generate a corresponding keypoint sequence \( \mathbf{K}_{\text{tar}} \) that follows the reference motion while maintaining the target's structural consistency.

\begin{comment}
For each frame \( t \), we estimate the global motion of the reference subject by fitting an ellipse to its keypoints, extracting its center \( C_{\text{ref}}^t \), orientation \( \theta_{\text{ref}}^t \), and scale factors along the major and minor axes. A similarity transformation is then applied to align the reference motion with the target subject:
\begin{equation}
    \hat{K}_{\text{tar}}^t = K_{\text{tar}}^1 + \mathbf{R}(\Delta \theta^t_\text{ref})\mathbf{S}(s_x, s_y) \Delta C^t_\text{ref}.
\end{equation}
Here, \( \mathbf{R}(\cdot) \) and \( \mathbf{S}(\cdot) \) denote rotation and scaling transformations, while \( s_x \) and \( s_y \) represent the scale factors extracted from the fitted ellipse.
\end{comment}

To further capture local movements for each keypoint, we model keypoint displacements in polar coordinates relative to the keypoint set center $O_{\text{tar}}^t$. Each keypoint's position is adjusted by a radial scaling factor and an polar angular shift computed from $K^t_\text{ref}$ and $K^0_\text{ref}$, ensuring that local motion variations are faithfully transferred.
\begin{comment}
\begin{equation}
    r_{\text{tar}}^{t,\text{new}} = s_r^t \cdot \|\hat{K}_{\text{tar}}^t - C_{\text{tar}}^t\|,\
    \phi_{\text{tar}}^{t,\text{new}} = \phi_{\text{tar}}^t + \Delta \phi_{\text{local}}^t.
\end{equation}
\end{comment}
The updated keypoints are then converted back into Cartesian coordinates and mapped to the global coordinate system.

\noindent{\textbf{TPS-based shape warping.}} % The target keypoint sequence $\mathbf{K}_\text{tar}$ sharing the similar motion with the reference keypoint sequence $\mathbf{K}_\text{ref}$. 
Naturally, $\mathbf{K}_\text{tar}$ can serve as a guiding option for video generation. However, we discovered that point-based guidance lacks continuity, which disrupts the temporal attention in our training-free video generation pipeline, resulting in undesirable outcomes. This finding motivates us to enhance high-level semantic motion alignment by integrating it with low-level morphological motion alignment. We utilize the correspondence established between $\mathbf{K}_\text{tar}$ and $\mathbf{K}_\text{ref}$ to reshape the reference object into the target shape by applying Thin Plate Spline (TPS) transformations~\cite{bookstein1989principal}. 
% convert the sparse keypoint sequence into dense shape sequence by applying Thin Plate Spline (TPS)\cite{bookstein1989principal} transformations per frame.

% we apply TPS transform on the source object frame by frame. Based on the correspondence between $\mathbf{K}_\text{ref}$ and $\mathbf{K}_\text{tar}$, TPS transformation will convert the source object into the target shape as shown in Figure~\ref{fig:shape-adaptation}.
Specifically, given \( \mathbf{K}_{\text{ref}}^t \) and \( \mathbf{K}_{\text{tar}}^t \), we estimate a warping function \( \mathcal{T}^t \) that satisfies~\cref{eq:warp}:
\begin{equation}\label{eq:warp}
  \mathbf{K}^t_{\text{tar}} = \mathcal{T}^t (\mathbf{K}^t_{\text{ref}})
\end{equation}
The TPS transformation $\mathcal{T}^t $  is parameterized as~\cref{eq:tps}:
\begin{equation}\label{eq:tps}
    \mathcal{T}^t (p) = A^t \begin{bmatrix} p \\ 1 \end{bmatrix} + \sum_{i=1}^{m} w^{t,i} \mathcal{U}(\|\mathbf{K}^{t,i}_{\text{tar}} - p\|^2),
\end{equation}
where $m$ is the number of keypoints, \( \mathcal{U}(r) = r^2 \log r^2 \) is a radial basis function, \( A^t \in \mathbb{R}^{2 \times 3} \) and \( w^{t,i} \in \mathbb{R}^{2 \times 1} \) are transformation parameters obtained by solving the bending energy~\cref{eq:energy},
\begin{equation}\label{eq:energy}
    \min_{\mathcal{T}^t} \int_{\mathbb{R}^2} \| H \|_F^2 \, dx \, dy.
\end{equation}
$\| H \|_F^2$ represents the Frobenius norm of the Hessian matrix, the second-order partial derivatives of $\mathcal{T}^t$ with respect to keypoint coordinates. 

\begin{figure*}[htbp]
\vspace{-1.5em}
  \centering
  \includegraphics[width=\textwidth]{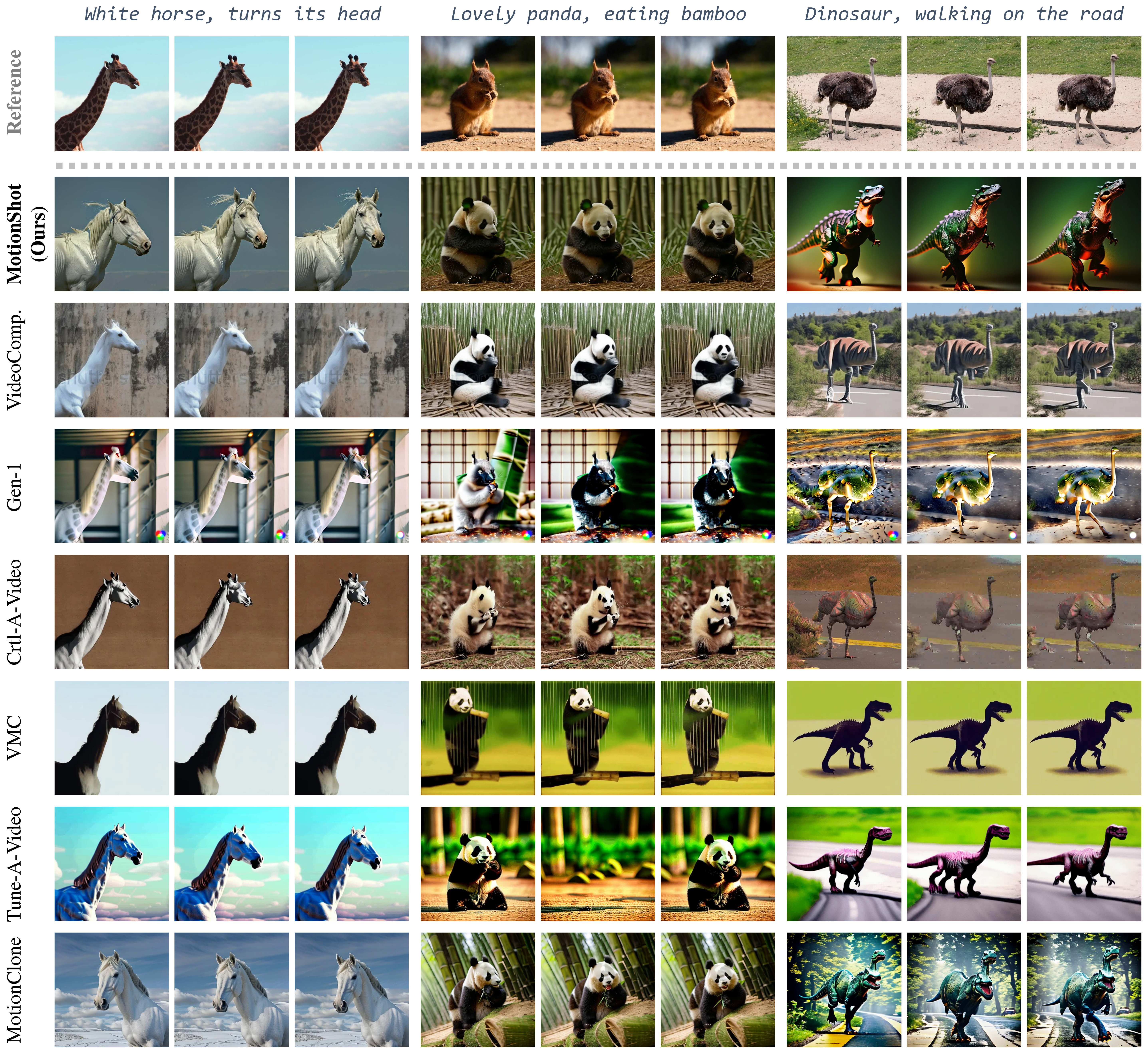}
  \vspace{-1.5em}
  \caption{\textbf{Visual comparison with baseline methods.} 
  \method demonstrates strong semantic alignment and excellent morphological accuracy, whereas baseline methods are influenced by the shape of the reference object, resulting in poor morphological outcomes (e.g., the horse's neck in the left column and the dinosaur's neck and paw in the right column).
  % \method generate a horse with natural shape while other methods lengthen the neck region according to the given giraffe reference, showing the lack of ability of morphological motion transfer.
  }
  \label{fig:main_compare}
  \vspace{-1.5em}
\end{figure*}

Finally, we warp reference video frames with estimated warp function $\mathcal{T}$ so as to obtain transformed reference object in the target shape while maintaining the original motion as shown in~\cref{fig:shape-adaptation}. By integrating semantic and morphological motion alignment, our approach effectively preserves both the motion of the reference object and the structure of the target object, enabling high-quality motion retargeting that aligns seamlessly with the reference video.

\subsection{Attention-guided Video Generation}\label{sec:3.4}
The TPS warped reference frames provide strong motion prior information for the video generation. We guide the video generation with the warped reference frames through scored-based function in a training-free manner. 

In this module, following~\cite{ling2024motionclone}, we first apply single-step noise addition and denoising operation to the warped frames to obtain the temporal attention map at a specific time step $\tau$, denoted as $A^{\tau}_{\text{ref}} \in \mathbb{R}^{(H \times W) \times C \times F \times F}$. Each element $[A^{\tau}_{\text{ref}}]_{p,i,j}$ captures the temporal correlation between frame $i$ and frame $j$ at spatial location $p$, satisfying the normalization constraint: \(\sum_{j=1}^{f} [A^{\tau}_{\text{ref}}]_{p,i,j} = 1\).

\begin{table*}[t]
\vspace{-1em}
    \centering
    \resizebox{0.9\textwidth}{!}{
    \tiny
    \begin{tabular}{lcc|cccc}
        \toprule
        & \multicolumn{2}{c|}{CLIP Scores $\uparrow$ } & \multicolumn{4}{c}{User Study $\uparrow$ }\\
        \cmidrule{2-7}
        & \makecell{Text \\ Alignment} & \makecell{Temporal \\ Consistency} 
        & \makecell{Motion \\ Preservation} & \makecell{Appearance \\ Diversity} 
        & \makecell{Text \\ Alignment} & \makecell{Temporal \\ Consistency} \\
        \midrule
        VideoComposer~\cite{wang2023videocomposer}       & 26.54 & 95.95 & 3.00 & 2.72 & 2.79 & 2.82 \\
        Gen-1~\cite{esser2023structure}               & 22.79 & 97.67 & 2.87 & 2.71 & 2.75 & 2.87 \\
        VMC~\cite{jeong2024vmc}                 & 26.77 & 97.72 & 2.80 & 2.78 & 2.78 & 2.87 \\
        Tune-A-Video~\cite{wu2023tune}        & 26.60 & 95.99 & 2.86 & 2.78 & 2.88 & 2.86 \\
        Control-A-Video~\cite{chen2023control}     & 24.87 & 95.54 & 2.94 & 2.66 & 2.40 & 2.92 \\
        MotionClone~\cite{ling2024motionclone}         & 26.41 & 97.48 & 2.90 & 2.50 & 2.80 & 2.82 \\
        \textbf{MotionShot (Ours)}       & \textbf{26.95} & \textbf{97.81} & \textbf{4.95} & \textbf{4.95} & \textbf{4.94} & \textbf{4.90} \\
        \bottomrule
    \end{tabular}
    }
    \vspace{-0.8em}
    \caption{\textbf{Quantitative comparison.} Our method significantly outperforms the other leading methods.}
    \label{tab:quantitative_evaluation}
    \vspace{-1.5em}
\end{table*}

Since $A^{\tau}_{\text{ref}}$ may contain noise and irrelevant information, to enhance the effectiveness of motion constraints, we select the top-k values along the temporal dimension for each frame. This to the construction of a sparse control mask $M^{\tau} \in \mathbb{R}^{(H \times W) \times C \times F \times F}$.

In the diffusion inference phase, the sampling process~\cite{dhariwal2021diffusion} can be guided by a customized energy function $g$ with guidance strength $\lambda$, enabling diffusion sampling to be conditioned on auxiliary information. To guide the generation, we define the energy function as~\cref{eq:sampling_energy}:
\begin{equation}\label{eq:sampling_energy}
g = \|M^{\tau} \cdot (A^{\tau}_{\text{ref}} - A^t_{\text{gen}}) \|_2^2
\end{equation}

Due to the warping of the reference frame sequence, the motion information in the temporal attention aligns with the structure of the target object. 
By integrating this into the diffusion model’s sampling process,
as \cref{eq:sample}, 
\begin{equation}
\hat{\epsilon}_{\theta} = \epsilon_{\theta}(z_t, \text{text}, t) - \lambda\nabla_{z_t} g(z_t; t, \text{reference video}),
\label{eq:sample}
\end{equation}
we impose constraints on the generated video's temporal attention map $A^t_{\text{gen}}$, ensuring its motion patterns closely align with the reference object's movement.

\section{Experiments}
\label{sec:experiment}

\subsection{Implement Details}
\label{sec:4.1}

In this work, we select AnimateDiff\cite{guo2023animatediff}, as the video generation framework.  
In the Semantic Motion Alignment module, we set the ControlNet condition weight to 0.6 and configure the control mode as \textit{`My prompt is more important'} to generate the fake target object.  
In the keypoint sampling operation, we set the interval $d$ as 200 in uniform contour sampling and sample total $m=30$ points. 
In the Attention-Guided Video Generation stage, following~\cite{ling2024motionclone}, we set the timestep $\tau$ to 400 and select $k = 1$. The primary attention map is extracted from the reference video within the first upsampling block of the U-Net.  
For the sampling process, we perform a total of 300 steps with the DDIM\cite{song2020denoising} scheduler, and the guidance is applied during the first 180 steps.

\vspace{-0.2em}
\subsection{Experiments Setup}
\noindent{\textbf{Dataset.}} Following~\cite{jeong2024vmc,ling2024motionclone}, our evaluation utilizes reference videos from the DAVIS dataset \cite{ponttuset20182017davischallengevideo} and various online resources, comprising a total of 40 videos. These videos encompass a diverse range of motion types exhibited by different subjects, including 10 videos featuring human motion, 20 videos capturing animal movement, and 10 videos depicting other dynamic scenes. % This diversity provides a comprehensive set of motion scenarios, enhancing the robustness of our evaluation.

\noindent{\textbf{Evaluation metrics.}} 
For objective evaluation, we adopt two widely recognized metrics from prior work \cite{jeong2024vmc, ling2024motionclone, guo2023animatediff}: \textit{textual alignment}, which measures how closely the generated video matches the given prompt, and \textit{temporal consistency}, which assesses the smoothness of motion. In addition to quantitative metrics, we conduct a user study to capture human judgment more comprehensively. A panel of 20 volunteers evaluates each approach, assigning scores from 1 to 5 based on four key aspects: \textit{motion preservation}, \textit{appearance diversity} between input and generated videos, and the \textit{text alignment} and \textit{temporal consistency} of the generated videos. The final score for each aspect is the average rating from the volunteers.

\vspace{-0.5em}
\subsection{Qualitative Results.} We compare our \method with state-of-the-art (SOTA) motion transfer methods as shown in~ \cref{fig:main_compare}, including VideoComposer\cite{wang2023videocomposer}, Gen-1\cite{esser2023structure}, VMC\cite{jeong2024vmc}, Tune-A-Video\cite{wu2023tune}, Control-A-Video\cite{chen2023control}, and MotionClone\cite{ling2024motionclone}. 
% Visual comparisons are presented in \cref{fig:main_compare}. 
VideoComposer, Gen-1, and Control-A-Video are constrained by the structure of the original video, making it challenging to generate a target object with natural shape.
Meanwhile, VMC and Tune-A-Video struggle to preserve the motion consistency of the original video, while MotionClone faces difficulties in ensuring compatibility between motion and appearance. 
In contrast, MotionShot effectively retargets motion information to align with the target subject, ensuring both natural motion dynamics and a coherent visual appearance in the generated video.

\vspace{-0.5em}
\subsection{Quantitative Results.} 
\cref{tab:quantitative_evaluation} 
presents a quantitative comparison based on CLIP scores and user study evaluations. \method achieves the highest scores for both text alignment and temporal consistency. Furthermore, in the user preference assessment, \method outperforms all baselines in all four aspects, demonstrating its strong capability in motion transfer.

\subsection{Ablation Study}
\label{sec:4.3}

\begin{figure}
    \centering
    \includegraphics[width=1.0\linewidth]{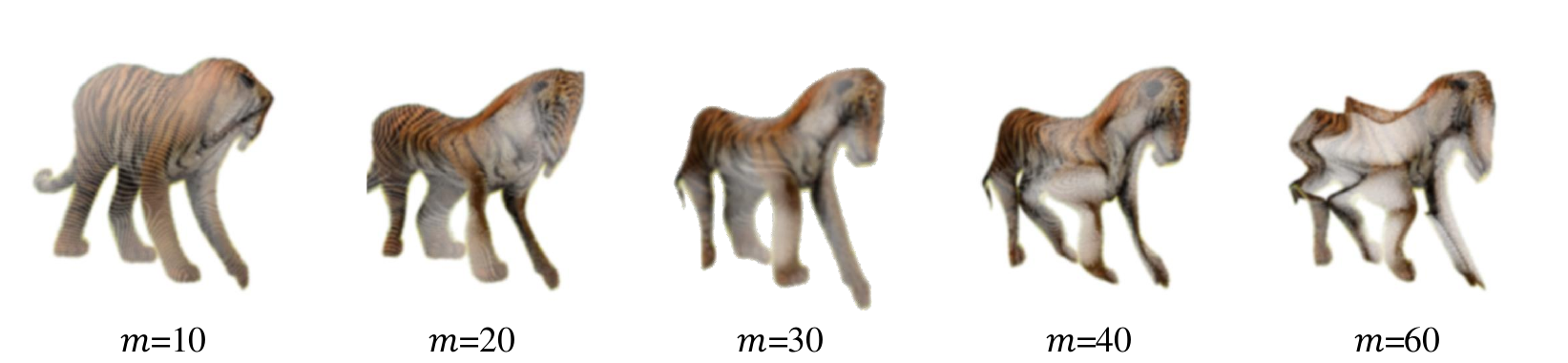}
    \vspace{-1.5em}
    \caption{\textbf{Ablation of number of sampled keypoints $m$.}}
    \label{fig:ablation_keypoints}
    \vspace{-1.5em}
\end{figure}

\begin{figure}[htbp]
\vspace{-1.5em}
  \centering
  \includegraphics[width=0.48\textwidth]{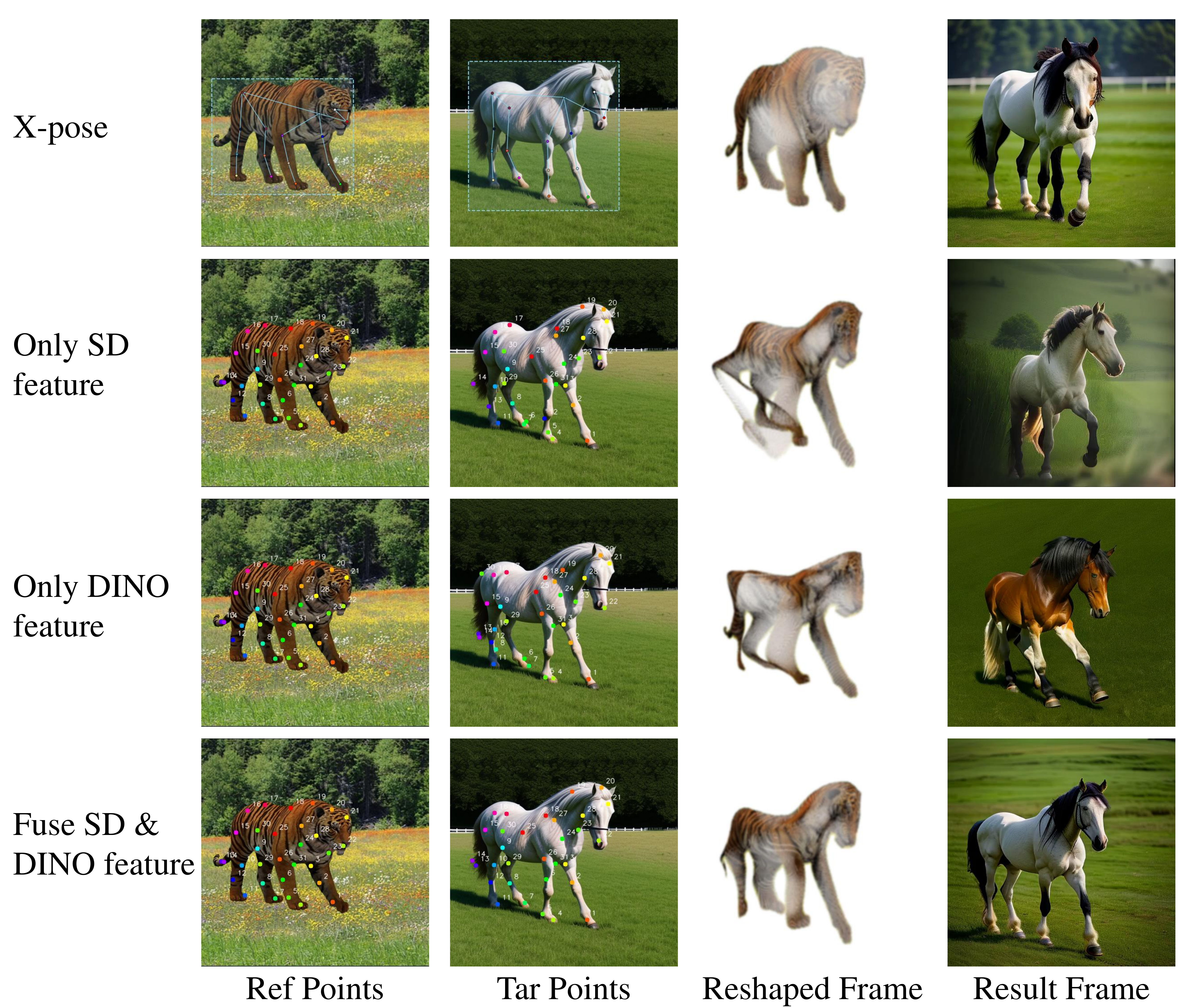}
  \vspace{-1.5em}
  \caption{\textbf{Influence of different keypoints matching methods.} Our proposed method (Fuse SD \& DINO feature) achieve best high-level semantic motion transfer result.}
  \label{fig:ablation_sematic}
  \vspace{-1em}
\end{figure}

\noindent{\textbf{Number of sampled keypoints.}}
In \cref{fig:ablation_keypoints}, we compare the impact of the number of sampled keypoints $m$. 
We proportionally adjust the number of sampled contour and internal points, ranging from $m=10$ (8 contour points, 2 internal points) to $m=60$ (48 contour points, 12 internal points). When $m$ is small (e.g.,$m=10$), the TPS transformation fails to deform the reference frame to match the target shape. Conversely, when the number of keypoints is too large (e.g., $m=60$), the deformation results exhibit overfitting. At $m=30$, the reference frame undergoes a reasonable deformation, making it our chosen value for all subsequent experiments.

\begin{figure}
  \centering
  \includegraphics[width=0.8\linewidth]{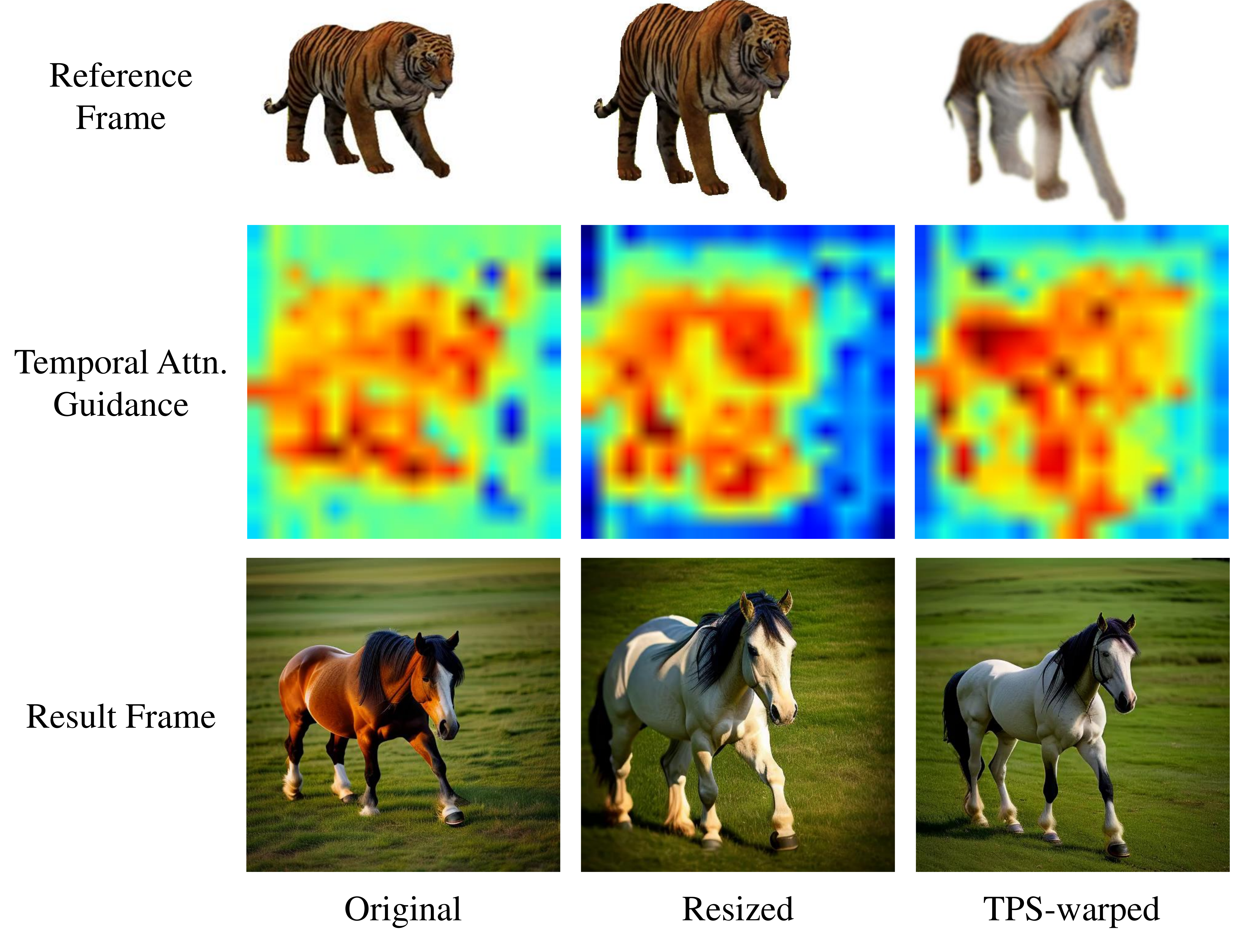}
  \vspace{-1.0em}
  \caption{\textbf{Influence of different shape retargeting methods.} Our method produces motion that is well-aligned with the target subject, resulting in more harmonious visual outcomes.}
  \label{fig:ablation_reshape}
  \vspace{-1.5em}
\end{figure}

\begin{figure}
\vspace{-1.5em}
    \centering
    \includegraphics[width=0.8\linewidth]{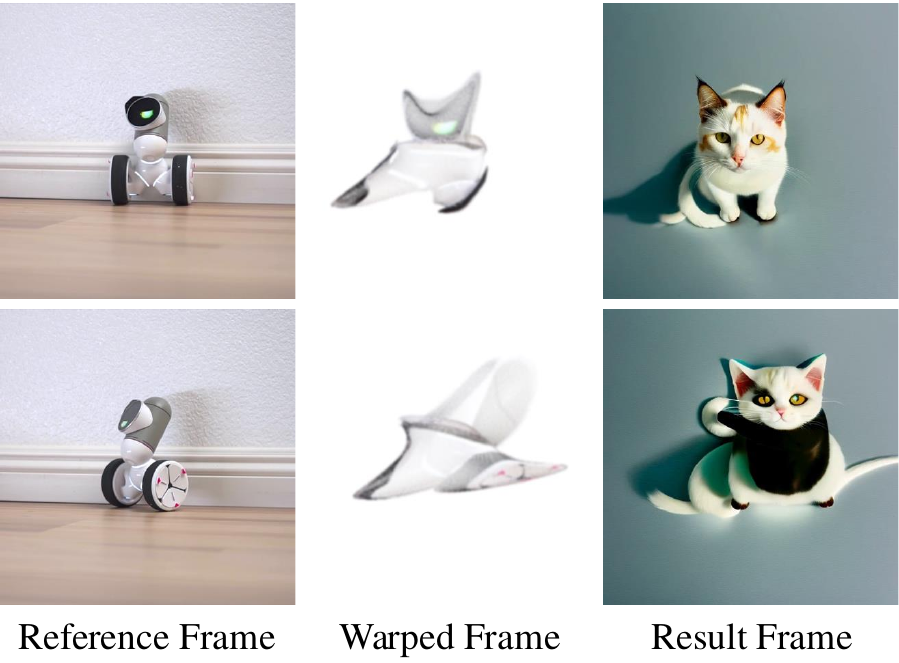}
    \caption{Limitation of \method, which will fail on reference-target pairs without any semantic similarities.}
    \label{fig:limitation}
    \vspace{-1.5em}
\end{figure}

\noindent{\textbf{Semantic feature matching.}} 
% To evaluate the effectiveness of our semantic feature matching for semantic motion alignment, we compare it with several optional methods. These include matching based on results from a pre-trained keypoint detector (e.g., X-Pose \cite{yang2024x}), as well as keypoint matching using only SD features or DINO features, as shown in \cref{fig:ablation_sematic}.
% Although the keypoint detector can accurately predict 17 key points for animals, these keypoints are not evenly distributed throughout the object, leading to discrepancies between the deformed appearance and the target appearance. 
% Since SD features primarily capture fine-grained spatial information, they are prone to matching errors in regions with ambiguous spatial details (e.g., the tail area in \cref{fig:ablation_sematic}), which can lead to inaccuracies in retargeting and transfer results.
% DINO features effectively capture high-level information from the input image but may produce errors when matching keypoints in fine spatial details (e.g., the horse's legs in \cref{fig:ablation_sematic}). 
% Our method combines SD and DINO features, effectively balancing fine-grained and high-level precision on semantic motion alignment.
% Furthermore, we adopt Poisson disk sampling to ensure an even distribution of keypoints across the subject, thereby enhancing the stability and consistency of the deformation.
To evaluate our semantic feature matching for motion alignment, we compare it with several methods, including matching from a pre-trained keypoint detector (e.g., X-Pose \cite{yang2024x}) and keypoint matching using only SD or DINO features, as illustrated in \cref{fig:ablation_sematic}.The keypoint detector predicts 17 landmarks for animals but suffers from uneven distribution, leading to appearance mismatches. SD features offer fine spatial detail but are error-prone in ambiguous areas (e.g., the tail), while DINO captures high-level semantics but may miss fine details (e.g., horse legs). Our method combines SD and DINO features to balance fine-grained and high-level precision in motion alignment. % Additionally, we use Poisson disk sampling to ensure an even distribution of keypoints, enhancing deformation stability and consistency.

% These matching errors can lead to distortions in the generated output during retargeting.
% Stable Diffusion (SD) features primarily capture low-level spatial information, ensuring spatial consistency in the absence of strong texture signals, while DINO features focus on high-level semantic information and excel in capturing sparse yet precise matching points. 
% Therefore, using SD or DINO features alone for keypoint matching results in varying degrees of matching errors, particularly in the limbs of animals, which ultimately leads to deformations in the subject. 

\vspace{0.5em}

\noindent{\textbf{TPS-based shape warping.}}
As shown in \cref{fig:ablation_reshape}, 
% red-highlighted areas in the temporal attention guidance indicate regions of significant motion amplitude that should align with the target object's shape (e.g., a horse) for accurate control.
regions with high motion amplitude (highlighted in red) should align with the target object's shape (e.g., a horse) for accurate control.
Using original sequences often results in motion-shape mismatches, distorting the generated horse's appearance to resemble a tiger (left column).% , as seen in the left column of \cref{fig:ablation_reshape}. 
Resizing improves size consistency but still introduces topological distortions, such as misaligned legs (middle column). In contrast, our keypoint-based retargeting preserves both motion accuracy and structural consistency.

% \section{Limitation}
% Our \method establishes both high-level semantic correspondence and low-level morphological correspondence between reference-target pairs using SD and DINO features, enabling coherent and precise motion transfer. It is important to note that \method operates effectively only when the reference-target objects share certain semantic similarities. 

\section{Limitation \& Conclusion}

\paragraph{Limitation. }
\method operates effectively mostly when reference-target objects share similar semantic. 
In cases of no similarities, \method may yield unpredictable results, as the semantic correspondence between the pairs cannot be correctly established, shown in~\cref{fig:limitation}. This is reasonable, as the model lacks any prior knowledge of motion alignment until additional cues are provided.

\paragraph{Conclusion.}
In this work, we present \method, a training-free motion transfer method that aligns both high-level semantic and low-level morphological motions.
To ensure semantic alignment, we propose a structure-aware keypoint sampling strategy and utilize fused SD and DINO features for semantic feature matching.
To address shape inconsistencies
% between the reference video and textual prompt
, we introduce a morphological motion alignment operation leveraging keypoint tracking and TPS transformation to warp objects % in the reference frames 
to the desired shapes.
% With the proposed methods, we obtain reference frames with the correct shape, enabling attention-guided motion transfer with strong semantic alignment and excellent morphological motion alignment.
These designs enable attention-guided motion transfer with strong semantic consistency and accurate shape alignment.
% Experimental results demonstrate that \method outperforms baseline methods in text alignment, temporal consistency, motion preservation, and appearance diversity.

\clearpage
\newpage

\noindent\textbf{Acknowledgments.}
This work was supported in part by the National Natural Science Foundation of China (Grant No. 62372133), in part by Shenzhen Fundamental Research Program (Grant NO. JCYJ20220818102415032).

{
    \small
    \bibliographystyle{ieeenat_fullname}
    \bibliography{main}
}

\end{document}